\documentclass{article}
\usepackage{ijcai24}

\usepackage{times}
\usepackage{soul}
\usepackage{url}
\usepackage[hidelinks]{hyperref}
\usepackage[utf8]{inputenc}
\usepackage[small]{caption}
\usepackage{graphicx}
\usepackage{amsmath}
\usepackage{amsthm}
\usepackage{booktabs}
\usepackage{algorithm}
\usepackage{algorithmic}
\usepackage[switch]{lineno}

\linenumbers

\title{Chatbot is Not All You Need: Information-rich Prompting for More Realistic Responses}

\author{
Seokhoon Jeong$^1$
\and
Assentay Makhmud$^2$\and
\affiliations
$^1$Ulsan National Institute of Science and Technology\\
$^2$Suleyman Demirel University
\emails
\{shjd0246, almacho\}@unist.ac.kr
}

\begin{document}
\nolinenumbers

\maketitle

\begin{abstract}
    Recent Large Language Models (LLMs) have shown remarkable capabilities in mimicking fictional characters or real humans in conversational settings. However, the realism and consistency of these responses can be further enhanced by providing richer information of the agent being mimicked. In this paper, we propose a novel approach to generate more realistic and consistent responses from LLMs, leveraging five senses, attributes, emotional states, relationship with the interlocutor, and memories. By incorporating these factors, we aim to increase the LLM's capacity for generating natural and realistic reactions in conversational exchanges. Through our research, we expect to contribute to the development of LLMs that demonstrate improved capabilities in mimicking fictional characters. We release a new benchmark dataset and all our codes, prompts, and sample results on our Github\footnote{\url{https://github.com/srafsasm/InfoRichBot}}.
\end{abstract}

\section{Introduction}
Large laguage models based on transformer architecture \cite{vaswani2017attention} have demonstrated impressive performance in a wide range of tasks \cite{brown2020language,devlin2018bert,radford2019language}. More recent LLMs that leverage both supervised fine-tuning and reinforcement learning from human feedback \cite{ouyang2022training} have shown their decent capabilities even in open-domain conversational tasks including human-like interactions \cite{liu2023summary}. However, due to the significant computational resources required for training such models, prompt tuning has emerged as a crucial aspect of optimizing LLM performance. Recent research has explored various techniques to generate more realistic responses through effective prompt engineering, such as prompting a relevant pseudo dialogue \cite{han2022meet} or providing detailed information of the scene, relations, and attributes \cite{chen2022would}. Despite these advances, several major challenges remain unaddressed, including: (1) LLMs mimicking fictional characters often exhibit either too little or too harsh fluctuations in their responses. (2) The innate context limit of LLMs poses a challenge for maintaining a consistent conversational memory. (3) A scarcity of conversational datasets containing helpful information to inform prompt construction.

To address these challenges, we present a multi-pronged approach aimed at enhancing the efficacy of prompts for LLMs:

1. \textbf{~Information-rich Prompting} We initialize and continuously update the prompts so that it provides multi-aspect information on the character. \\\\
2. \textbf{~Within-prompt Self Memory Management} To mitigate the limitation of context length, we make the language model to summarize the history log and maintain it in the prompt. \\\\
3.\textbf{~Benchmark Dataset} To overcome the scarcity of useful datasets for evaluation, we augment Cornell Movie-Dialog Corpus\footnote{\url{https://www.cs.cornell.edu/~cristian/Cornell_Movie-Dialogs_Corpus.html}} via GPT-3.5 Turbo, a model known for its strong capabilities comparable to those of fine-tuned LLMs \cite{zhong2023can}.\\
In this paper, we detail the methodology and results of our proposed techniques, demonstrating their potential to address the challenges currently faced in LLM prompt engineering.

\section{Related Work}
\subsection{Realistic Dialogue Generation}
Although recent LLMs have demonstrated decent performance in terms of general conversation, it is still challenging to generate a realistic response, mimicking a fictional character or a real human. There has been an attempt to reflect the attributes or personalities assigned to the chatbot \cite{li2020aloha}, or even those of the interlocutor \cite{fernau2022towards,liu2020you,chen2022would}. Some others tried to bring external knowledge in addition to persona information \cite{lim2023you}, while another leveraged pseudo dialogue of the mimicked character \cite{han2022meet}. Such trials have demonstrated partial successes leading to more emotional or realistic conversation.

\subsection{Prompt Engineering for LLMs}
Prompt engineering has emerged as a practical alternative to address the substantial cost of training LLMs. Three popular types of prompts are widely used: template-based prompts, few-shot prompts, and chain-of-thought prompts. Structural prompts involve creating templates or patterns in the input query that guide the model to produce desired responses, such as providing information in a key-value structure \cite{zhong2022proqa}. These patterns help the model understand the structure and context of the task, making it more likely to generate coherent and relevant answers. Few-shot prompting is an approach where the model is first provided with a small number of examples or demonstrations of a specific task before being asked to perform the task itself \cite{brown2020language}. Chain-of-thought prompting tries to elicit LLMs' reasoning capacity by providing prompts to solve a problem step-by-step \cite{wei2022chain}.

\subsection{Text Emotion Recognition}
Text emotion recognition is a task that aims to extract emotional information from natural language. Traditional machine-learning based approaches mostly focused on extracting lexical, syntatic, and semantic features from the natural language text \cite{alm2005emotions,strapparava2008learning} or additional features such as n-grams, part-of-speech tags, and word clusters \cite{aman2007identifying} to achieve the task. More recent researches take advantage of deep learning based architectures like Recurrent Neural Networks \cite{medsker2001recurrent}, Long Short-Term Memory \cite{graves2012long}, or Transformer \cite{vaswani2017attention}. Leveraging transformer-based language models or combining them with RNN or LSTM have shown state-of-the-art results in emotion recognition or classification tasks \cite{li2020hierarchical,wang2020contextualized}. Very recently, it has been also claimed that LLMs trained using both supervised fine-tuning and reinforcement learning with human feedback demonstrates comparable zero-shot performance to fine-tuned BERT-style models \cite{zhong2023can}.

\section{Approach}

\begin{figure}
  \centering
  \includegraphics[width=8cm,height=3cm]{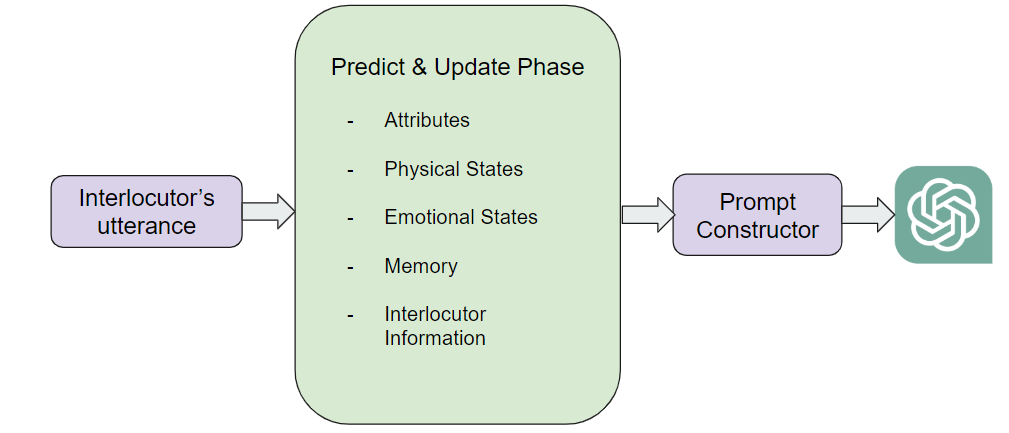}
  \caption{Structure of information-rich prompted generation.}
\end{figure}

\subsection{Information-rich Prompting}
Information-rich prompting provides the language model with comprehensive information about the characters involved in the conversation prior to generating the next utterance. The provided information includes attributes of the characters, their physical states, emotional states, memories, and knowledge about their interlocutors.

The attributes of the characters encompass a wide range of characteristics such as their names, personality traits, and personal goals. These attributes are not explicitly defined and can be generated by the language model from an open domain, allowing for flexibility and adaptability in character portrayal.

To incorporate the characters' physical states, we enable the model to perceive their sensory experiences, including sight, hearing, taste, smell, and touch. These physical states are updated after each interlocutor's utterance, taking into account the impact of the interlocutor's words and actions on the character's senses.

Emotional states play a crucial role in shaping the characters' responses and behavior throughout the conversation. Therefore, we ensure that the language model predicts how the characters' emotional states might change as a result of the interlocutor's utterance. Given the complex and multifaceted nature of emotions, we maintain a list of several emotions to accurately represent the characters' emotional landscapes.

In order to facilitate coherent and contextually appropriate conversations, we incorporate the characters' memories. These memories serve as a summarized log of the ongoing conversation or past experiences, allowing the model to refer back to previous interactions and maintain continuity in the dialogue.

Additionally, the characters possess knowledge about their interlocutors, which is divided into three main components. Intuitively, the utterance may become different due to different relationship, favorability, or experiences with the interlocutor. This information does not include an objective and comprehensive information on the interlocutor, but only reflects the awareness of the character. This seems to be more realistic since it is impossible to know everything about the interlocutor. 

In summary, the "Information-rich Prompting" technique enriches the conversational abilities of the language model by providing a comprehensive set of inputs. By incorporating characters' attributes, physical states, emotional states, memories, and knowledge on interlocutors, we create a more nuanced and contextually aware conversational AI system, capable of generating coherent and engaging dialogues.
Knowledge on interlocutors consists of three parts. First is relationship, which represents the character's relationship with the interlocutor. 

\subsection{Self Memory Management}
In the context of maintaining the memories of a chatbot, our approach involves two essential components: the conversation log and memory management. The conversation log can be regarded as the chatbot's short-term memory, while the memory list serves as its long-term memory. This division allows for efficient storage and retrieval of information.

The conversation log is given as multiple lines of strings, wherein both the chatbot's own utterances and actions, as well as those of the interlocutor model, are recorded. This real-time log captures the ongoing dialogue and serves as a temporary repository of recent interactions. To prevent unbounded growth, we set a predetermined threshold for the length of the conversation log.

When the conversation log reaches this threshold, a summarization process is initiated. The language model is instructed to condense the contents of the conversation log into a concise one-line summary. This summary encapsulates the key points and highlights of the conversation. The generated summary is then appended to the memory list, effectively transitioning the information from short-term to long-term storage.

By implementing this approach, we ensure that the chatbot possesses both short-term and long-term memory capabilities. The conversation log enables real-time context awareness, while the memory management allows for the retention of important information over extended periods. This structured memory framework enhances the chatbot's ability to engage in meaningful and coherent conversations.

\section{Experiments}
\subsection{Data}
Regarding the datasets for generating more realistic conversation, numerous datasets have been proposed. Yet, these datasets primarily focus on only one factor among the emotional state during an utterance \cite{li2017dailydialog}, the personality of the speaker \cite{li2020aloha}, or the relationship between agents \cite{urbanek2019learning}. To the best of our knowledge, no dataset comprehensively incorporates information regarding characters' personalities, their relationship with interlocutors, and the emotional states during the utterances. To address this gap, we present an augmented version of a small portion of Movie Dialog Corpus, which we will refer to as the Dialogue-Emotion-Attributes-Relationship (DEAR) dataset.

The original Movie Dialog Corpus contains 9,035 characters from 617 movies, and a total of 220,579 conversational exchanges between these characters, amounting to 304,713 utterances. We selected this corpus due to its diverse range of dialogues, emotions, and relationships depicted in the movies, and the fact that we can easily gather information related to movie characters from an online website, Fandom\footnote{\url{https://www.fandom.com/}}. Fandom is an online website which includes information on a variety of creations, written by general users.

Although Fandom is a massive database, not every movie and characters' information is available because most of the movies in Movie Dialog Corpus are before 2000 and the utterances from trivial characters are also included. Our dataset consists of 7,514 sets of conversations carefully selected from a diverse range of 113 movies. To enrich this dataset with additional information, we leveraged the capabilities of the GPT-3.5-turbo language model. Although this model's performance may not be considered exceptionally outstanding, it offers distinct advantages in generating character attributes and emotions within an open domain, unlike other models that rely on fixed attribute sets. Given the nuanced nature of human attributes and emotions, we anticipated that the resulting dataset would exhibit a wide array of diverse cases.

In addition to the movie dialogues themselves, our dataset incorporates several crucial elements. First, we provide attributes for the characters involved in each conversation, ensuring a comprehensive understanding of their individual characteristics. Second, we include the sensory information of each character. Third, a summary of background information preceding each conversation offers contextual details to aid in comprehension and analysis. Finally, we capture the emotional state of the characters, providing insights into their mental and psychological states during the dialogue exchanges.

\subsection{Evaluation Method}
To evaluate the quality of our prompt engineering approach, we plan to utilize a combination of automatic and human evaluation methods on DEAR. For automatic evaluation, we employed a well-established metrics such as Metric for Evaluation of Translation with Explicit Ordering (METEOR) \cite{banerjee2005meteor} and Sentence BERT \cite{reimers2019sentence}, which have been widely used in the field of natural language processing to assess the quality of generated text. A detailed context for each situation will be given as a prompt, and our prompt-engineering-guided output would be compared to the ground truth outputs using aforementioned metrics.

\subsection{Experimental Details}
In this section, we present the experimental procedure employed to evaluate the performance of our approach, utilizing the GPT3.5-Turbo language model. For each conversation, we let the language model to summarize 200 tokens of the original movie script that appear before the conversation. We acknowledge that is much reliable and robust to use more tokens for the background information, but this choice was made due to the financial constraints. Then, based on the information gathered from Fandom and the script, the language model generates an utterance in a conversation.

\subsection{Results}
We demonstrate that each component in our approach helps generating a better utterance. $GPT_{raw}$ means raw gpt, only given with the previous interlocutor's utterance and the character's information. $GPT_{sense}$, $GPT_{emotion}$, $GPT_{memory}$, $GPT_{interlocutor}$ each stands for generation with an additional component in the subscript, and $GPT_{full}$ is a case that incorporates all the components. We show in \autoref{evaluation-results}, that combining the components appraently helps generating a better utterance.

\begin{table}
  \caption{Evaluation Results}
  \label{evaluation-results}
  \centering
  \begin{tabular}{lll}
    \toprule
    Model     & METEOR     & Sentence BERT \\
    \midrule
    $GPT_{raw}$     & $0.113$ & $0.208$      \\
    $GPT_{sense}$     & $0.112$       & $0.205$  \\
    $GPT_{emotion}$     & $0.105$       & $0.201$  \\
    $GPT_{memory}$     & $0.116$       & $0.206$  \\
    $GPT_{interlocutor}$     & $0.114$       & $0.204$  \\
    $GPT_{full}$     & $\mathbf{0.118}$       &  $\mathbf{0.209}$  \\
    \bottomrule
  \end{tabular}
\end{table}

\section{Analysis}
For each case, giving an additional information does not necessarily help the language model to generate more coherent and realistic utterance. This seems to be the limitation of DEAR dataset, since it includes erroneous data collected from unintended website. Nevertheless, combining them improves the ability to generate utterances. Further analysis of the evaluation dataset would be required.

\section{Conclusion}
In conclusion, our study highlights the effectiveness of information-rich prompting in enhancing the naturalness and realism of utterance generation when the language model emulates a fictional character. 

However, it is important to acknowledge the limitations that exist in our current work. First, DEAR dataset includes some unreliable information derived from two cases: erroneous information retrieval from Fandom and hallucinations of language models. We observed some of such cases in our dataset, so it requires a careful inspection. Second, a comparative analysis with other state-of-the-art utterance generation models has not been conducted, and it would be valuable to assess the performance of our approach against alternative methodologies. Third, no human evaluation was done due to time and financial constraints. Results in \autoref{evaluation-results} do not demonstrate high scores, but due to the fact that the metrics only check the similarity to the ground truth, predictions that might be better than the ground truth would have been overlooked. Further evaluation by humans is expected to mitigate such side-effects. Fourth, ethical considerations have not been thoroughly examined. During preliminary investigations, we observed that our approach significantly weakens the ethical filtering of the gpt3.5-turbo model, potentially leading to the generation of inappropriate content such as violence, sexuality, or anti-social behaviors. Consequently, a comprehensive ethical analysis from multiple perspectives is imperative. Future research should address these limitations to provide a more comprehensive understanding of the effectiveness, generalizability, and ethical implications of our information-rich prompting approach.

\section{Acknowledgement}
This research was conducted during 2023.03.02.-2023.06.16. as a course project on UNIST CSE402 Natural Language Processing. We thank for professor Taehwan Kim for advising our work. This is an ongoing work and may be updated without a notice.

\bibliographystyle{named}
\bibliography{ijcai24}

\end{document}